\documentclass[11pt,a4paper]{article}
\usepackage[hyperref]{emnlp2019}
\usepackage{times}
\usepackage{latexsym}

\usepackage[utf8]{inputenc}
\usepackage{color,soul}
\setul{0.45ex}{0.25ex}

\usepackage{makecell}
\usepackage{graphicx}
\usepackage{subcaption}

\usepackage{amsmath}
\usepackage{booktabs}
\usepackage{multirow}
\usepackage{graphicx}
\usepackage{array}
\usepackage{pgfplots}
\usepackage{algorithm}
\usepackage{url}

\usepackage{listings}
\usepackage{xcolor}
\usepackage{times}
\usepackage{latexsym}
\usepackage{amsmath}
\usepackage{algorithm}
\usepackage{amsfonts}
\usepackage[noend]{algpseudocode}
\usepackage{lipsum}
\usepackage{multirow}
\usepackage{framed}
\usepackage{graphicx}
\usepackage{longtable}
\usepackage{tikz}
\usepackage{booktabs}
\usepackage[all]{nowidow}
\usepackage{soul}
\usepackage{spverbatim}
\usepackage{enumitem}
\usepackage{bm}
\aclfinalcopy 

\newcommand{\name}{AllenNLP Interpret}

\usepackage{amssymb}

\title{AllenNLP Interpret:\\A Framework for Explaining Predictions of NLP Models}

\def\authorSpace{\hspace{4mm}}
\author{\makecell{Eric Wallace$^1$ \authorSpace{} Jens Tuyls$^2$ \authorSpace{} Junlin Wang$^2$ \authorSpace{} Sanjay Subramanian$^1$\\
Matt Gardner$^1$ \authorSpace{} Sameer Singh$^2$} \\
$^1$Allen Institute for Artificial Intelligence\authorSpace{} 
$^2$University of California, Irvine\\
\href{mailto:ericw@allenai.org}{\tt ericw@allenai.org}, \href{mailto:sameer@uci.edu}{\tt sameer@uci.edu}
}

\date{}

\begin{document}
\maketitle
\begin{abstract}
Neural NLP models are increasingly accurate but are imperfect and opaque---they break in counterintuitive ways and leave end users puzzled at their behavior. 
Model interpretation methods ameliorate this opacity by providing explanations for specific model predictions. 
Unfortunately, existing interpretation codebases make it difficult to apply these methods to new models and tasks, which hinders adoption for practitioners and burdens interpretability researchers. 
We introduce AllenNLP Interpret, a flexible framework for interpreting NLP models. The toolkit provides interpretation primitives (e.g., input gradients) for any AllenNLP model and task, a suite of built-in interpretation methods, and a library of front-end visualization components.
We demonstrate the toolkit's flexibility and utility by implementing live demos for five interpretation methods (e.g., saliency maps and adversarial attacks) on a variety of models and tasks (e.g., masked language modeling using BERT and reading comprehension using BiDAF). These demos, alongside our code and tutorials, are available at \url{https://allennlp.org/interpret}.

\end{abstract}
  
\section{Introduction}

\begin{figure}
\includegraphics[trim={0.75cm 6.6cm 9cm 0.5cm},clip, width=1.4\columnwidth]{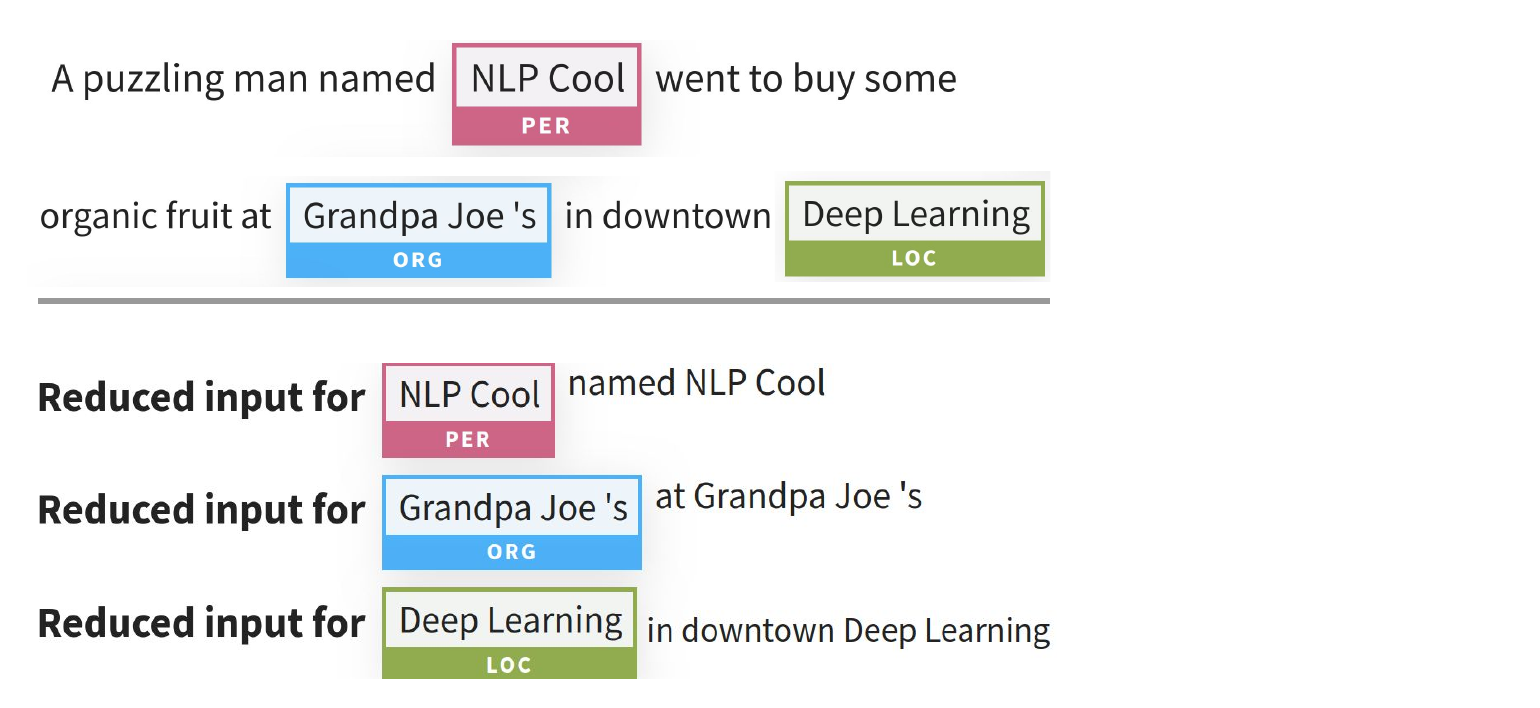}
\caption{An interpretation generated using \name{} for NER. The model predicts three tags for an input (top). We interpret each tag separately, e.g., input reduction~\cite{feng2018pathologies} (bottom) removes as many words as possible without changing a tag's prediction. Input reduction shows that the words ``named'', ``at'', and  ``in downtown'' are sufficient to predict the \texttt{People}, \texttt{Organization}, and \texttt{Location} tags, respectively.}
\label{fig:ner}
\end{figure}

Despite constant advances and seemingly super-human performance on constrained domains, state-of-the-art models for NLP are imperfect: they latch on to superficial patterns~\cite{gururangan2018annotation}, reflect unwanted social biases~\cite{doshi2017towards}, and significantly underperform humans on a myriad of tasks. These imperfections, coupled with today's advances being driven by (seemingly black-box) neural models, leave researchers and practitioners scratching their heads, asking, ``\emph{why did my model make this prediction?}''

Instance-level interpretation methods help to answer this question by providing explanations for specific model predictions. These explanations come in many flavors, e.g., visualizing a model's local decision boundary~\cite{ribeiro2016should}, highlighting the saliency of the input features~\cite{simonyan2013saliency}, or adversarially modifying the input~\cite{ebrahimi2017hotflip}. Interpretations are useful to illuminate the strengths and weaknesses of a model~\cite{feng2018pathologies}, increase user trust~\cite{ribeiro2016should}, and evaluate hard-to-define criteria such as safety or fairness~\cite{doshi2017towards}.

Many open-source implementations exist for instance-level interpretation methods. However, most codebases focus on computer vision, are model- or task-specific (e.g., sentiment analysis), or contain implementations for a small number of interpretation methods. Thus, it is difficult for practitioners to interpret \emph{their} model. As a result, model developers rarely leverage interpretations and thus lack a robust understanding of their system. The inflexibility of existing interpretation codebases also burdens interpretability researchers---they cannot easily evaluate their methods on multiple models. 

We present \name{}, an open-source, extensible toolkit built on top of AllenNLP~\cite{garnder2018allen} for interpreting NLP models. The toolkit makes it easy to apply existing interpretation methods to \emph{new models}, as well as develop \emph{new interpretation methods}. The toolkit consists of three contributions: a suite of interpretation techniques implemented for broad classes of models, model- and task-agnostic APIs for developing new interpretation methods (e.g., APIs to obtain input gradients), and reusable front-end components for interactively visualizing the interpretations.

\name{} has numerous \textbf{use cases}. Our external website shows demos of:
\begin{itemize}[nosep,leftmargin=3mm]
    \item \emph{Uncovering Model Biases:} A SQuAD model relies on lexical overlap between the words in the question and the passage. Alternatively, a textual entailment model infers contradiction on observing the word ``politics'' in the hypothesis.
    \item \emph{Finding Decision Rules:} A named entity recognition model predicts the location tag when it sees the phrase ``in downtown''.
    \item \emph{Diagnosing Errors:} A sentiment model incorrectly predicts the positive class due to the trigram ``tony hawk style''.
\end{itemize}
\section{Interpreting Model Predictions}\label{sec:background}

\begin{figure*}[t]
\centering
\includegraphics[trim={0cm 0cm 0cm 0cm},clip, width=0.9\textwidth]{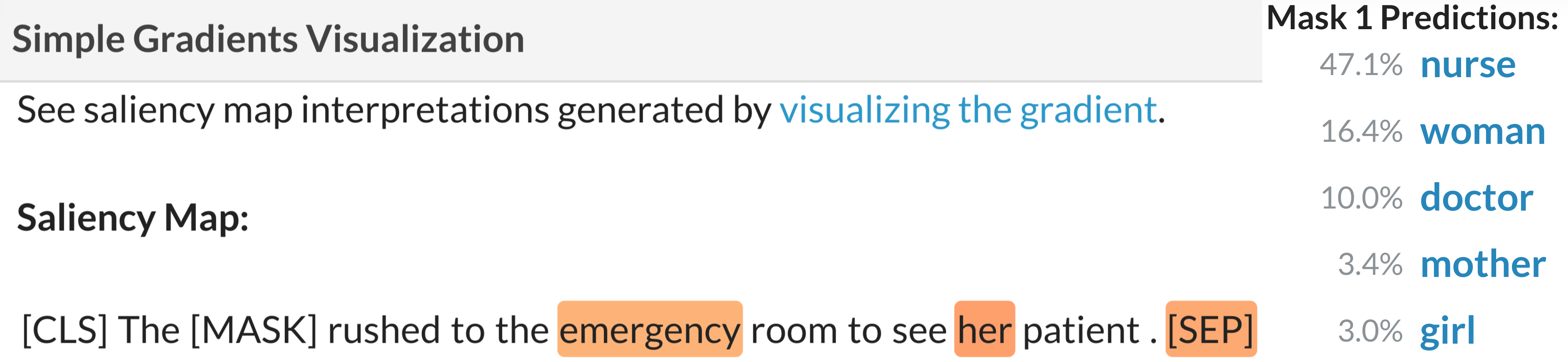}
\caption{A saliency map generated using Vanilla Gradient~\cite{simonyan2013saliency} for BERT's masked language modeling objective. BERT predicts the \texttt{[MASK]} token given the input sentence; the interpretation shows that BERT uses the gendered pronoun ``her'' and the hospital-specific ``emergency'' to predict ``nurse''.}
\label{fig:bert}
\end{figure*}

This section introduces an end user's view of our toolkit, i.e., the available interpretations, models, and visualizations.

\subsection{What Are Instance-Level Interpretations} \name{} focuses on two types of interpretations: gradient-based saliency maps and adversarial attacks. We choose these methods for their flexibility---gradient-based methods can be applied to any differentiable model.

Saliency maps explain a model's prediction by identifying the importance of the input tokens. Gradient-based methods determine this importance using the gradient of the loss with respect to the tokens~\cite{simonyan2013saliency}.

Adversarial attacks provide a different lens into a model---they elucidate its capabilities by exploiting its weaknesses. We focus on methods that modify tokens in the input (e.g., replace or remove tokens) in order to change the model's output in a desired manner.

\subsection{Saliency Map Visualizations}\label{subsec:saliency} 

We consider three saliency methods. Since our goal is to interpret why the model made \emph{its} prediction (not the ground-truth answer), we use the model's own output in the loss calculation. For each method, we reduce each token's gradient (which is the same dimension as the token embedding) to a single value by taking the $L_2$ norm.

\paragraph{Vanilla Gradient} This method visualizes the gradient of the loss with respect to each token~\cite{simonyan2013saliency}. Figure~\ref{fig:bert} shows an example interpretation of BERT~\cite{devlin2018BERT}.

\paragraph{Integrated Gradients} \citet{sundararajan2017axiomatic} introduce integrated gradients. They define a baseline $\bm{x}^\prime$, which is an input absent of information (we use a sequence of all zero embeddings). Word importance is determined by integrating the gradient along the path from this baseline to the original input. 

\paragraph{SmoothGrad} \citet{smilkov2017smoothgrad} average the gradient over many noisy versions of the input. For NLP, we add small Gaussian noise to every embedding and take the average gradient value.

\subsection{Adversarial Attacks}\label{subsec:attacks}

We consider two adversarial attacks: replacing words to change the model's prediction (HotFlip) and removing words to maintain the model's prediction (Input Reduction).

\paragraph{Untargeted \& Targeted HotFlip} We consider word-level substitutions using HotFlip~\cite{ebrahimi2017hotflip}. HotFlip uses the gradient to swap out words from the input in order to change the model's prediction. It answers a sensitivity question: \emph{how would the prediction change if certain words are replaced?} 
We also extend HotFlip to a targeted setting, i.e., we substitute words in order to change the model's prediction to a \emph{specific} target prediction. This answers an almost counterfactual question: \emph{what words should be swapped in order to cause a specific prediction?} 

\begin{figure*}[t]
\centering
\includegraphics[trim={0.7cm 5.6cm 0.7cm 0.6cm},clip, width=0.9\textwidth]{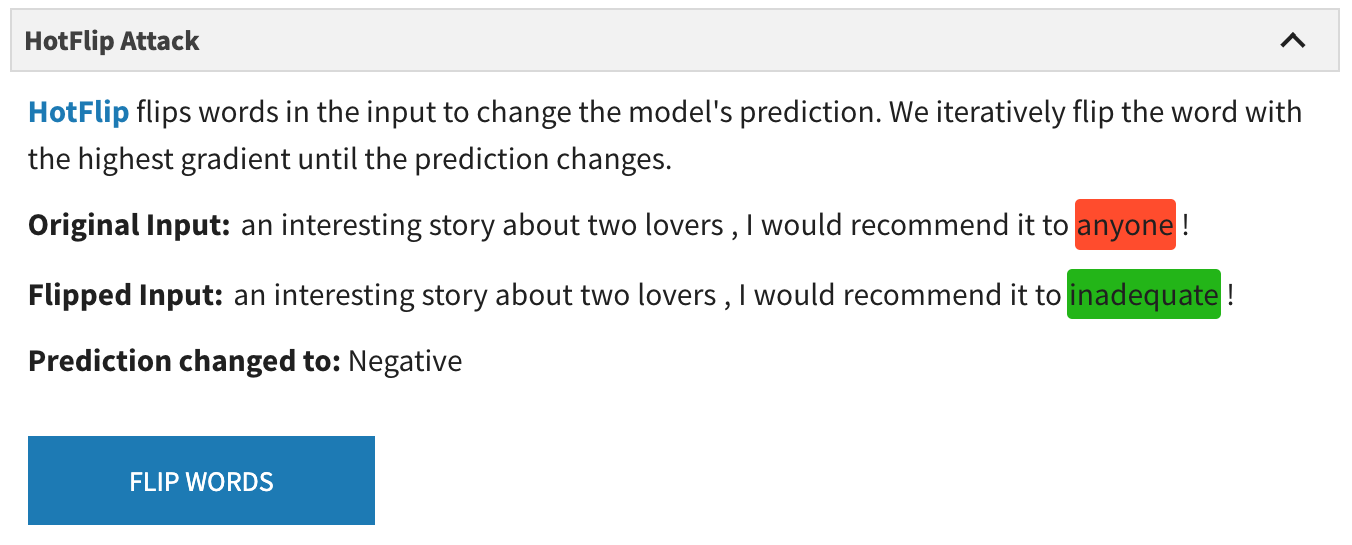}
\caption{A word-level HotFlip attack on a sentiment analysis model---replacing ``anyone'' with ``inadequate'' causes the model's prediction to change from Positive to Negative.}
\label{fig:hotflip}
\end{figure*}

We closely follow the original HotFlip algorithm: replace tokens based on a first-order Taylor approximation of the loss around the current token embeddings.\footnote{We also adapt HotFlip to contextual embeddings; details provided in Section~\ref{subsec:contextual}.}
Figure~\ref{fig:hotflip} shows an example of a HotFlip attack on sentiment analysis.

\paragraph{Input Reduction} \citet{feng2018pathologies} introduce input reduction. They remove as many words as possible from the input \emph{without} changing a model's prediction. Input reduction works by iteratively removing the word with the smallest gradient value. We classify input reduction as an ``adversarial attack'' because the resulting inputs are usually nonsensical but cause high confidence predictions~\cite{feng2018pathologies}. Figure~\ref{fig:ner} shows an example of reducing an NER input.

\subsection{Currently Available Models}

The toolkit currently interprets six tasks which cover a wide range of input-output formats and model architectures.

\begin{itemize}[leftmargin=4mm]
    \itemsep 0pt
    \item \textbf{Reading Comprehension} using the SQuAD~\cite{rajpurkar2016squad} and DROP~\cite{drop} datasets. We use NAQANet~\cite{drop} and BiDAF models~\cite{Seo2017BidirectionalAF}.
    \item \textbf{Masked Language Modeling} using the transformer models available in \texttt{Pytorch Transformers}\footnote{\href{https://github.com/huggingface/pytorch-transformers}{https://github.com/huggingface/pytorch-transformers}}, e.g., BERT~\cite{devlin2018BERT}, RoBERTa~\cite{liu2019roberta}, and more. \item \textbf{Text Classification} and \textbf{Textual Entailment} using BiLSTM and self-attention classifiers.
    \item \textbf{Named Entity Recognition (NER)} and \textbf{Coreference Resolution}. These are examples of tasks with complex input-output structure; we can use the same function calls to analyze each predicted tag (e.g., Figure~\ref{fig:ner}) or cluster.
\end{itemize}
\section{AllenNLP Interpret Under the Hood}\label{sec:implementation}

\begin{figure*}[tbh]
\centering
\includegraphics[trim={0cm 5.5cm 0cm 0cm},clip, width=\textwidth]{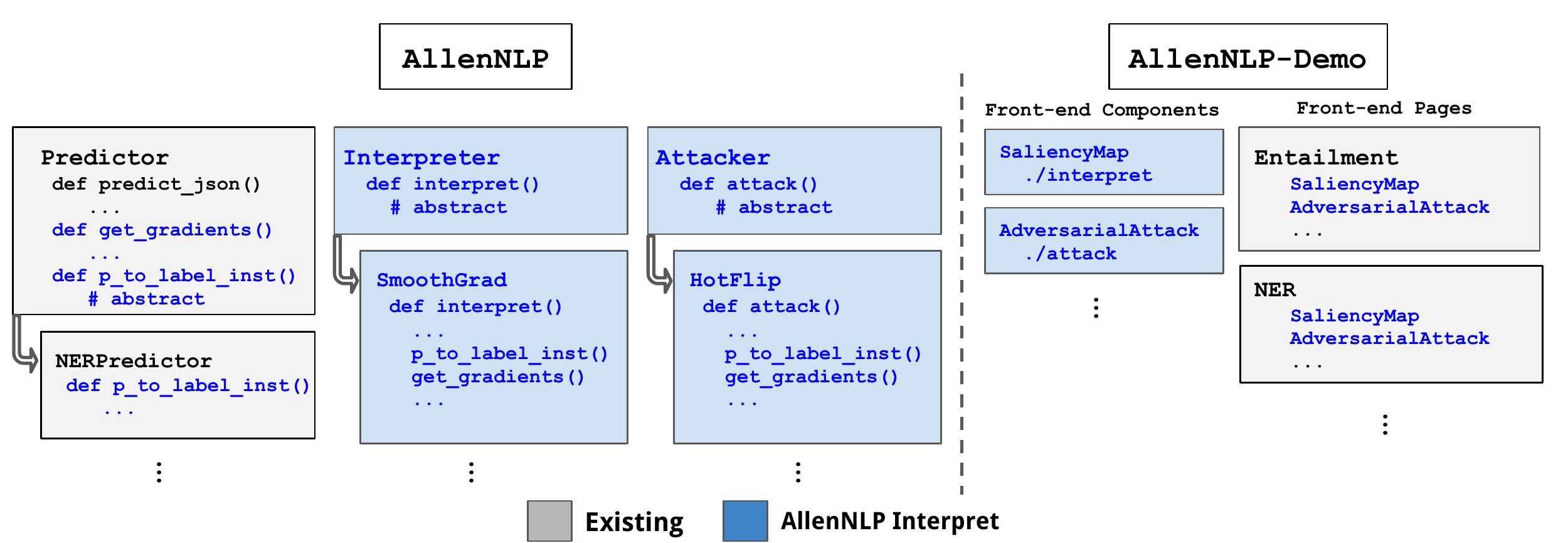}
\caption{\textbf{System Overview:} Our toolkit (in \textcolor{blue}{blue}) and the surrounding AllenNLP ecosystem. The only model-specific code is a simple function called \texttt{predictions\_to\_labeled\_instances()} (abbreviated as \texttt{p\_to\_label\_inst()}), which is added to the model's \texttt{Predictor} class (e.g., for an NER model's predictor; left of figure). This function allows input gradients to be calculated using \texttt{get\_gradients()} in a model-agnostic manner (e.g., for use in SmoothGrad or HotFlip; middle left of Figure). On the front-end (right of Figure), we create reusable visualization components, e.g., for visualizing saliency maps or adversarial attacks.}
\label{fig:software}
\end{figure*}

This section provides implementation details for \name{}: how we compute the token embedding gradient in a model-agnostic way, as well as the available front-end interface. Figure~\ref{fig:software} provides an overview of our software implementation and the surrounding AllenNLP ecosystem.

\subsection{Model-Agnostic Input Gradients}

\paragraph{Existing Classes in AllenNLP} Models in AllenNLP are of type \texttt{Model} (a thin wrapper around a PyTorch \texttt{Module}). The \texttt{Model} wrapper includes a \texttt{forward()} function, which runs the model and optionally computes the loss if a label is provided.

Obtaining predictions from an AllenNLP \texttt{Model} is simplified via the \texttt{Predictor} class. This class provides a model-agnostic way for obtaining predictions: call \texttt{predict\_json()} with a JSON containing raw strings and it will return the model's prediction. For example, passing \texttt{\{``input'': ``this demo is amazing!''}\texttt{\}} to a sentiment analysis \texttt{Predictor} will receive positive and negative class probabilities in return.

\paragraph{Our AllenNLP Extension} The core backbone of our toolkit is an extension to the \texttt{Predictor} class that allows interpretation methods to compute input gradients in a model-agnostic way. Creating this extension has two main implementation challenges: (1) the loss (with the model's \emph{own predictions} as the labels) must be computed for widely varying output formats (e.g., classification, tagging, or language modeling), and (2) the gradient of this loss with respect to the token embeddings must be computed for widely varying embedding types (e.g., word vectors, ELMo~\cite{PetersELMo2018} embeddings, BERT embeddings). 

\paragraph{Predictions to Labeled Instances} To handle challenge (1), we leverage the fact that all models will return a loss if a label is passed to their \texttt{forward()} function. We first query the model with the input to obtain its prediction. 
Next, we convert this prediction into a set of ``labeled examples'' using a function called \texttt{predictions\_to\_labeled\_instances()}. For categorical predictions (e.g., classification, span prediction), this function returns a single instance with the label set to the model's argmax prediction. 

For tasks with structured outputs (e.g., NER, coref), this function returns multiple instances, where each instance is used to compute the loss for a different part of the output. For example, there are separate instances for each of the three NER tags predicted in Figure~\ref{fig:ner}. Separating out the instances allows us to have more fine-grained interpretations---we can analyze one part of the overall prediction rather than interpreting the entire tag sequence.

\paragraph{Embedding-Agnostic Gradients} To handle difficulty (2)---computing the gradients of varying token embeddings---we rely on the abstractions of AllenNLP. In particular, AllenNLP uses a \texttt{TokenEmbedder} interface to converts token ids into embeddings. We can thus compute the gradient for any embedding method by registering a PyTorch backward gradient hook on the model's \texttt{TokenEmbedder} function. 

Our end result is a simple API for computing input gradients for any model: call \texttt{predictions\_to\_labeled\_instances()} and then \texttt{get\_gradients()}.

\subsection{Context-Independent Embedding Matrix for Deep Embeddings}\label{subsec:contextual}

The final implementation difficulty arises from the fact that contextual embeddings such as ELMo and BERT do not have an ``embedding matrix'' to search over (their embeddings are context-dependent). This raises difficulties for methods such as Hotflip (Section~\ref{subsec:attacks}) that require searching over a discrete embedding matrix. To solve this, we create a context-independent matrix that contains the features from the model's last context-independent layer. For instance, we pass all of the words from a particular task's training set into ELMo and save the features from its context-independent Char-CNN into a ``word embedding matrix''. This allows us to run HotFlip for contextual embeddings while still capturing context information since the gradient backpropagates through the contextual layers.

\subsection{Frontend Visualizations} We interactively visualize the interpretations using the \href{https://demo.allennlp.org}{AllenNLP Demo}, a web application for running AllenNLP models. We add HTML and JavaScript components that provide visualizations for saliency maps and adversarial attacks. These components are reusable and greatly simplify the process for adding new models and interpretation methods (Section~\ref{sec:new}). For example, a single line of HTML code can create the visualizations shown in Figures 1--3.
Note that visualizing the interpretations is not required---AllenNLP Interpret can be run in an offline, batch manner. This is useful for aggregating interpretation results, e.g., as in \citet{feng2018pathologies} and \citet{wallace2018Neighbors}.

\section{Adding a Model or Interpretation}\label{sec:new}

This section describes the high-level process for adding new analysis methods or AllenNLP models to our toolkit. 

\paragraph{New Interpretation} We provide a tutorial for adding a new analysis method to our toolkit. In particular, it walks through the three main requirements for adding SmoothGrad:
\begin{enumerate}[nosep]
   \itemsep 0pt
    \item Implementing SmoothGrad in AllenNLP, using \texttt{predictions\_to\_labeled\_instances()} and \texttt{get\_gradients()} (requires adding about ten lines of code to the vanilla gradient method).
    \item Adding a SmoothGrad Interpreter to the demo back-end (about five lines of code).
    \item Adding the HTML/JavaScript for saliency visualization (requires making a one-line call to the reusable front-end components).
\end{enumerate}

\paragraph{New Model}
We also provide a tutorial for interpreting a new model. If your task is already available in the demos (e.g., text classification), you need to change a \emph{single} line of code to replace the demo model with your model. 
If your task is not present in the demos, you will need to:
\begin{enumerate}[nosep]
   \itemsep 0pt
    \item Write the \texttt{predictions\_to\_labeled\_instances()} function for your model (consists of three lines for classification).
    \item Create a path to your model in the demo's back-end (about 5-10 lines of code).
    \item Add a front-end page to visualize the model and interpretation output. This is simplified by the reusable front-end components (consists of copy-pasting code templates).
\end{enumerate}
\section{Related Work}

\textbf{Alternative Interpretation Methods} We focus on gradient-based methods (saliency maps and adversarial attacks) but numerous other instance-level model interpretation methods exist. For example, a common practice in NLP is to visualize attention weights~\cite{bahdanau2014neural} or to isolate the effect of individual neurons~\cite{karpathy2016visualizing}. We focus on gradient-based methods because they are applicable to many models.~\smallskip 
    
\noindent \textbf{Existing Interpretation Toolkits} In computer vision, various open-source toolkits exist for explaining and attacking models (e.g., \citet{papernot2016technical,uozbulak_pytorch_vis_2019}, inter alia); some toolkits also include interactive demos~\citep{norton2017adversarial}. Similar toolkits for NLP are significantly scarcer, and most toolkits focus on specific models or tasks. For instance, \citet{liu2018visual}, \citet{strobelt2019seq}, and \citet{vig2019visualizing} visualize attention weights for specific NLP models, while \citet{lee2018qadiver} apply adversarial attacks to reading comprehension systems. Our toolkit differs because it is flexible and diverse; we can interpret and attack any AllenNLP model.
\section{Conclusion}

We presented \name{}, an open-source toolkit that facilitates the interpretation of NLP models. The toolkit is flexible---it enables the development and evaluation of interpretation methods across a wide range of NLP models and tasks. 

The toolkit is continually evolving---we will continue to implement new interpretation methods and models as they become available. We welcome open-source contributions, and we hope the toolkit is useful for model developers and interpretability researchers alike.

\section*{Acknowledgements}

The authors thank Shi Feng, the members of UCI NLP, and the anonymous reviewers for their valuable feedback. We also thank the developers of AllenNLP for their help with constructing our toolkit, especially Joel Grus. This work is supported in part by NSF Grant IIS-1756023. 

\bibliography{journal-abbrv,bib}
\bibliographystyle{acl_natbib}

\end{document}